\documentclass[fullpaper,cameraready]{nldl}

\renewcommand{\DOI}{10.7557/18.6798}

\usepackage[utf8]{inputenc}
\usepackage{url}
\usepackage{graphicx}
\usepackage{authblk}
\usepackage[colorlinks=true,citecolor=violet,anchorcolor=purple]{hyperref}\usepackage{url}
\usepackage{multicol}
\usepackage{multirow}
\usepackage{algorithm}
\usepackage{algorithmic}
\usepackage{booktabs}
\usepackage{minitoc}
\usepackage{comment}
\usepackage{subcaption}
\usepackage{amsmath}
\usepackage[subtle]{savetrees}
\usepackage{amsfonts}
\usepackage[square,numbers]{natbib}
\usepackage{balance}

\def \Cbf{{\mathbf C}}

\def \Ibf{{\mathbf I}}

\def \0bf{{\mathbf 0}}

\def \Lcal{{\mathcal L}}

\let\bb\mathbb
\let\c\mathcal

\DeclareMathOperator{\bern}{Bern}
\def\similarity{s}

\newcommand{\set}[1]{\lbrace#1\rbrace}

\def\prob{\bb P}

\let\paragraphold\paragraph{}
\renewcommand{\paragraph}[1]{\vspace{-1em}\paragraphold{#1}}

\title{Efficient Self-Supervision using Patch-based Contrastive Learning for Histopathology Image Segmentation}

\author[1]{Nicklas Boserup}
\author[1,2]{Raghavendra Selvan\thanks{Corresponding Author: raghav@di.ku.dk}}
\affil[1]{Dept. of Computer Science, University of Copenhagen}
\affil[2]{Dept. of Neuroscience, University of Copenhagen}
\date{\vspace{-5ex}}

\begin{document}
\nldlmaketitle
\doparttoc 
\faketableofcontents 

\part{} 
\maketitle
\begin{abstract}  
Learning discriminative representations of unlabelled data is a challenging task. Contrastive self-supervised learning provides a framework to learn meaningful representations using learned notions of similarity measures from simple pretext tasks. In this work, we propose a simple and efficient framework for self-supervised image segmentation using contrastive learning on image patches, without using explicit pretext tasks or any further labeled fine-tuning. A fully convolutional neural network (FCNN) is trained in a self-supervised manner to discern features in the input images and obtain confidence maps which capture the network's belief about the objects belonging to the same class. Positive- and negative- patches are sampled based on the average entropy in the confidence maps for contrastive learning. Convergence is assumed  when the information separation between the positive patches is small, and the positive-negative pairs is large. The proposed model only consists of a simple FCNN with 10.8k parameters and requires about 5 minutes to converge on the high resolution microscopy datasets, which is orders of magnitude smaller than the relevant self-supervised methods to attain similar performance. We evaluate the proposed method for the task of segmenting nuclei from two histopathology datasets, and show comparable performance with relevant self-supervised and supervised methods. \footnote{Source code available at: \url{https://github.com/nickeopti/bach-contrastive-segmentation}}
\end{abstract}

\section{Introduction}
Learning task specific representations without any -- or with limited -- labelled data continues to be an elusive goal of machine learning. Recent advancements in contrastive learning and self-supervised learning have shown promising results in obtaining discriminative representations of the data which can be useful for downstream applications such as image classification~\cite{chen2020simple}, object detection~\cite{xie2021unsupervised} and speech recognition~\cite{baevski2020wav2vec}. Contrastive self-supervised learning (CSL) has been successfully used as a form of pre-training to reduce the dependence on labeled data for more complex tasks such as image segmentation~\cite{xie2020instance}. Most self-supervised methods rely on pretext tasks for training them~\cite{noroozi2016unsupervised,sahasrabudhe2020self}. Designing relevant pretext tasks can be challenging and even if a useful pretext task is obtained they may not easily generalise across datasets~\cite{misra2020self}.

\begin{figure*}[t]
    \centering
    \includegraphics[width=0.95\textwidth]{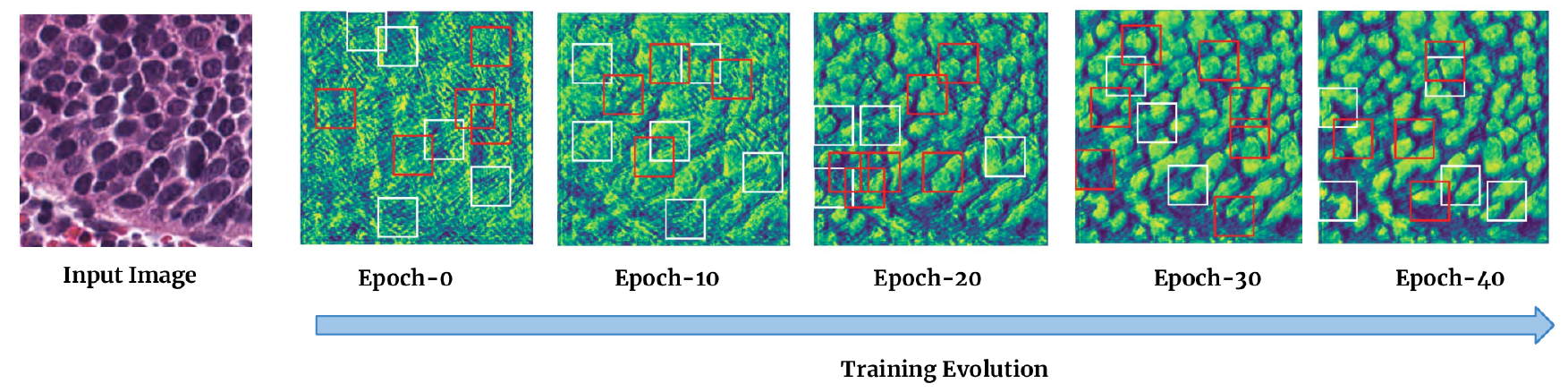}
    \caption{Overview of the training evolution of the proposed contrastive self-supervised learning model. 
    At each of the five illustrated epochs, predicted confidence map (from which segmentation masks are derived by thresholding) for a single validation set image is shown, along with the patches sampled as positive (white squares) and negative (red squares).}
    \label{fig:evolution}
    \vspace{-0.5cm}
\end{figure*}

In this work, we present a self-supervised learning framework that contrastively learns an object detection model for segmenting nuclei in histopathology images. The model comprises a fully convolutional neural network (FCNN) that predicts one confidence map per output channel, which captures the confidence of each pixel belonging to a particular object class. The FCNN is contrastively trained using smaller positive- and negative patches stochastically sampled from the images. Patches within a training batch are sampled from an entropy based distribution, where the entropy is based on the patch-level confidence scores. The intuition behind the entropy-based sampling is to obtain positive patches that contain similar information and negative patches that contain contrasting information, with respect to features that can discriminate between objects of different classes. Through iterative training we are able to improve the information separation between the positive- and negative patches, resulting in an object detection model which can be used for segmentation.

We use a simple FCNN with 10.8k tunable parameters which converges in about 5 minutes on a stand-alone GPU workstation. Experimental evaluation on two histopathology datasets \cite{monuseg_data,graham2019hover} show that our efficient, contrastive self-supervised learning method obtains performance comparable to relevant supervised and self-supervised baseline methods, using only a fraction of the compute time and resources. An illustration of how the confidence map evolves during the training process is illustrated in Figure~\ref{fig:evolution}.


\begin{figure*}[t]
    \centering
    \includegraphics[width=\textwidth]{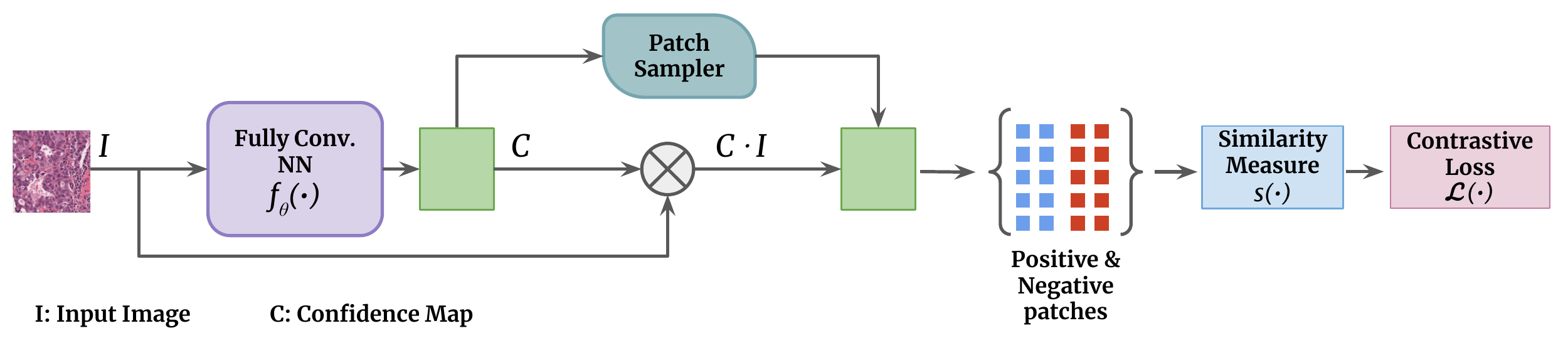}
    \caption{High level overview of the proposed patch-based contrastive self-supervision method. The input color image, $\Ibf$, is input to the fully convolutional neural network, $f_\theta(\cdot)$, to obtain a pixel level foreground confidence map, $\Cbf$. The confidence map is used by the {\em patch sampler} which returns locations of positive and negative patches. The patches are then obtained from the attended image, $\Cbf \cdot \Ibf$, to obtain the inter- and intra- class similarity measures, $s(\cdot)$. A contrastive loss, $\Lcal(\cdot)$, is computed based on these similarities which is then back-propagated through the entire pipeline. The trainable weights, $\theta$, are tuned until the confidence map, $\Cbf$, corresponds to segmentation masks of interest. }
    \label{fig:design}
\end{figure*}

\section{Method} 


Segmentation is fundamentally the task of partitioning an image into areas of interest and a background class. Assuming that the areas of interest within the same class are similar in some feature space according to a {\em similarity measure}, we present our contrastive learning framework that results in an unsupervised segmentation model.  A high level overview of the proposed framework is illustrated in Figure~\ref{fig:design}. 

\subsection{{Notation}}
\label{sec:notation}

Consider a batch of $M$ images, with the $i$'th image represented as the pair $(U_i, I_i)$, where $U_i \subset \bb N^d$ is a finite set representing the $d$-dimensional pixel coordinates and $I_i$ is a function $I_i \colon U_i \to [0,1]^\lambda$, with $\lambda \in \bb N$ denoting the number of colours/channels, mapping such pixel coordinates into their respective values. 

Next, we introduce the notation for denoting the patch locations sampled from the image $i$ as the tuple $(R,i)$, where $R \subset U_i$ is the set of pixel coordinates of the image patch. Note that the power set $\mathcal{P}(U_i)$ denotes the set of all possible smaller patches that can be sampled from image $i$. Denoting the set of all such patch tuples across images $\mathcal X$, 
we consider the subset $S \subset \mathcal X$ of regular (square) and spatially connected patches in this work.



The proposed framework uses a \emph{confidence network} (Section~\ref{sec:conf_net}), $f_\theta$, parameterised by $\theta$ which for a given input image $(U_i, I_i)$ computes the \emph{confidence map}, $C_i$, i.e., $f_\theta: (U_i,I_i) \in [0,1]^\lambda  \to C_i \in [0,1]^K$.
The $k$'th coordinate in $C_i^k(u)$ indicates the belief, or confidence, that the pixel $u \in U_i$ in image $i$ belongs to class $k$, for $k = 1,\dots,K$, with $K \geq 1$. 

The contrasting of patches depends on a similarity measure (Section~\ref{sec:sim}), $\similarity \colon \c X \times \c X \to \bb R $, which compares the similarity of two patches. This similarity computation relies on $(I_i, I_j)$ and $(C_i, C_j)$, where $i$ and $j$ are the images from which the two compared patches originate from.

\subsection{Confidence Network}
\label{sec:conf_net}
The aim of the confidence network, $f_\theta$, is to learn to detect objects in images  {without any pixel-level supervision}, such that areas with high confidence of the $k$'th class actually belong to a specific type of object in the images. Conversely, objects detected with high confidence of the $j$'th class, when $k \neq j$, should belong to a different class.  The confidence network, $f_\theta$, can be any trainable model which given an input image produces a set of \emph{confidence maps}, describing the confidence that each pixel of the image belongs to a corresponding class.
This, for instance, can be implemented as a fully convolutional neural network (FCNN), with multi-class support i.e., with multiple output channels such that each channel would correspond to a different output class. 

The confidence network is trained to discriminate between objects that could belong to different classes by contrasting \emph{patches} of images against each other. Obtaining meaningful positive- and negative patches for the contrastive learning is the foundational problem in this framework. We next describe a novel approach to mine for such contrastive patches for self-supervision of the confidence network.



\subsection{Entropy-based Patch Sampler}
\label{sec:sampling}


For each output class $k = 1,\dots,K$, patches {\em believed} to contain (a part of) an object belonging to a specific class are treated as \emph{positive} patches, whereas patches {\em believed} to not contain (a part of) an object of that class are treated as \emph{negative} patches. Appropriate sampling of such positive- and negative patches is of vital importance, as the optimisation of the confidence network, $f_\theta$, is performed according to the contrastive loss computed based on those patches.
Improved object detection shall ideally correlate with positive and negative sampled patches being more distinct. 
Effectively, the appropriate sampling of patches becomes a form of pretext/auxiliary task for training the confidence network, $f_\theta$. 

%
%

Assume a set of candidate patches, $S \subset \c X$. For each patch $(R, i) \in S$ with $|R|$ pixels and for each class $k$, the average patch confidence is computed as:
\begin{equation}
   A_k(R, i) := \frac{1}{|R|} \sum_{u \in R} C_i^k(u) \in [0, 1]. 
\end{equation}
Notice that the higher the average patch confidence, the stronger the belief that the patch contains the type of object belonging to class $k$. 

Recall that $C_i$ has range $[0, 1]^K$, that is, for each class, $k = 1,\dots,K$, a $C_i^k(u)$ value closer to 1 indicates a higher probability of $u$ belonging to the $k$'th class. Likewise, a value closer to 0 indicates that $u$ likely does not belong the $k$'th class, whereas values in between indicate varying degrees of (un)certainty in either direction. Using this intuition about the confidence maps, we model a Bernoulli distributed random variable, $X(u)$, with confidence value as its parameter, 
denoted as $X(u) \sim \bern(C_i^k(u))$. Using this, we can now define the average patch entropy as
\begin{equation}\small
   B_k(R, i) := \frac{1}{|R|} \sum_{u \in R} H(X(u))
\end{equation}
where
\begin{equation}\small
    H(X) = - \sum_{x \in \set{0, 1}} \prob(X(u) = x) \log_2 \prob(X(u) = x)
    \end{equation}
    \begin{equation}\small
    = -C_i^k(u) \log_2 C_i^k(u) - (1-C_i^k(u)) \log_2(1 - C_i^k(u)),
\end{equation}
which is the entropy of the Bernoulli distributed random variable $X(u)$. The motivation  for this choice of Bernoulli random variable is so that good choices of positive- and negative patches, according to the contrastive loss, would correlate with higher certainty from the confidence network. 
Therefore sampling a set $W_k \subset S$ of $n$ patches according to the unnormalised distribution
\[ 1 - B_k(R, i) \]
for each class $k$ ensures a stochastic correlation between the confidence map and the appropriateness of patch sampling. Finally,  partitioning of these $W_k$ into sets of positive samples, $P_k$ and negative samples $N_k$ is performed according to
\[ [ (R, i) \in P_k ] \sim \bern(A_k(R, i)) \]
for each patch $(R, i) \in W_k$, such that patches with high confidence of belonging to class $k$ are likely to be selected as positive for class $k$. 



\subsection{Similarity Measures}
\label{sec:sim}

Ideally, the similarity measure, $\similarity \colon \c X \times \c X \to \bb R$, should yield higher values when two input patches are more similar. In this work, for any two patches $(R_1, i), (R_2, j) \in S$, the similarity measure performs the comparison on the image scaled by the corresponding confidence map, given by the following pixel-wise products
\begin{align}
    F_i(u) &:= I_i(u) C_i(u), \quad u \in R_1 \\
    F_j(v) &:= I_j(v) C_j(v), \quad v \in R_2
\end{align}
We employ two pixel-level\footnotemark{} similarity measures; mean squared error, or difference, (MSE) and mean cross-entropy (CE), and treat them as a model hyperparameter and report the best performing measures for each of the datasets.
We envision density based similarity measures to be desirable, but in our limited experiments they performed poorly.
\footnotetext{Patches are compared pixel-by-pixel; formally, given a bijection $g$ between elements $u \in R_1$ and $v \in R_2$, patches are compared by $\similarity(F_i(u), F_j(g(v)))$. In practice, we compare pixels at same relative positions within the equisized patches.}
%
\paragraph{Backpropagation:} The scaling of the input image with the confidence map also serves the purpose of connecting the gradients between the sampling step and the confidence network for backpropagating the contrastive loss to optimise the confidence network.

\subsection{Contrastive Loss}
Within a channel $k$ of the confidence map, we seek to maximise similarity between positive patches while minimising the similarity between positive- and negative- patches. This is facilitated using the intra-channel contrastive loss
\begin{equation}\small
   \Lcal_\text{intra} = - \sum_{k = 1}^K \frac{1}{|P_k|^2}\sum_{r,p \in P_k} \log \frac{\exp(\similarity(r, p))}{\sum_{n \in N_k} \exp(\similarity(r, n))}.
\end{equation}
This contrastive loss, inspired by SimCLR \cite{chen2020simple}, rewards $f_\theta$ for learning to detect a single feature in the images for each class in the confidence map, when patches are sampled appropriately. However, this lacks any mechanism for enforcing the individual classes to learn distinct features. In a multi-class scenario, each of the classes shall ideally detect different features. This is achieved by introducing an inter-channel contrastive loss
\begin{equation}\small
   \Lcal_\text{inter} = -\sum_{k = 1}^K \frac{1}{|P_k|^2} \sum_{r,p_k \in P_k}  \frac{\exp(\similarity(r, p_k))}{\sum_{i = 1,[i \neq k]}^K  \sum_{p \in P_i} \exp(\similarity(r, p))}. 
\end{equation}
%
Finally, the combined loss is defined as
\begin{equation}
   \Lcal = \Lcal_\text{intra} + \Lcal_\text{inter}. 
\end{equation}



\section{Data \& Experiments}

\subsection{Data} \label{sec:data}

As the proposed CSL framework is trained exclusively using unlabelled images, without exposure to the labels during training, we can also use the labels from the training set for validation of the segmentation performance. That is, we train the weights of the FCNN without using any labels but can employ the labels for model selection. 
\paragraph{MoNuSeg:}
This dataset~\cite{monuseg_data}\footnote{MoNuSeg data is available at \url{https://monuseg.grand-challenge.org/Data/}} from MICCAI 2018 contains 37 training images at $1000 \times 1000$ pixels resolution, obtained at $40\times$ magnification, as well as 14 testing images at the same resolution. Dense annotations of all nuclei in all images are provided. 
\paragraph{CoNSeP:}
This dataset \cite{graham2019hover}\footnote{CoNSeP data is available at \url{https://warwick.ac.uk/fac/cross_fac/tia/data/hovernet/}} contains 27 training images at $1000 \times 1000$ pixels resolution, obtained at $40\times$ magnification, as well as 14 similar testing images. There are dense annotations of all nuclei in all images.





\subsection{Experimental Set-up}
\label{sec:baseline}
\paragraphold{Baseline Methods:} The simplest supervised baseline method is to obtain the optimal intensity threshold using the training images. The images are converted to grey-scale and an optimal threshold to obtain binary segmentation is obtained. Additionally, a CNN of the same architecture as the confidence network, $f_\theta$, is used as a supervised baseline. And finally, these results are also compared to the self-supervised method that uses scale prediction as a pretext \cite{sahasrabudhe2020self}, which is referred to as the Scale Pretext method. All baselines are evaluated on both mentioned datasets.
%
%
\paragraph{Model Hyperparameters:}
\label{sec:hyper} 
The default configuration uses a confidence network, $f_\theta$, closely inspired by \cite{sahasrabudhe2020self}. 
Entropy based stochastic sampling described in Section~\ref{sec:sampling} and $K = 4$ classes are used. The patch size of $50 \times 50$ pixels and 10 patches from each image in a batch are chosen based on experiments on the MoNuSeg dataset, as reported in Table~\ref{tab:hyperparam}. 

\begin{table}
\centering
\scriptsize
\begin{tabular}{@{}l|lllll@{}}
\toprule
\# Patches    & 5      & {\bf 10}     & 15     & 20     & 25      \\ 
Dice & 0.6643 & 0.7071 & 0.6682 & 0.6986 & 0.6059  \\ \midrule
Patch Size      & 20$^2$   & 30$^2$  & {\bf 50$^2$}  & 80$^2$  & 120$^2$\\
Dice & 0.5063 & 0.6562 & 0.6980 & 0.6606 & 0.6871  \\ \bottomrule
\end{tabular}
\caption{The median validation Dice score over three runs on MoNuSeg is used to select  number of patches and patch sizes (shown in boldface).}
\label{tab:hyperparam}
\end{table}

The proposed framework was trained for a maximum of 300 epochs. To limit RAM usage, the input images are cropped into $300 \times 300$ pixels, where the location is selected uniformly at random as to add some data variance. No other pre-processing nor data augmentation is applied.  Batch size of 10 is used for all experiments. 
\paragraph{Implementation:} The proposed framework is implemented in PyTorch \cite{pytorch}, with support to be trained on GPUs; all training has been performed on a system with an NVIDIA GeForce RTX 3060 and intel-i7 processor with 32GB memory. 
%
\paragraph{Experiments:} With the objective of comparing the segmentation performance of the proposed CSL framework with the baselines, we perform experiments on each of the datasets described in Section~\ref{sec:data}. 

Our CSL framework is initialised with the hyperparameters described in Section~\ref{sec:hyper} on each of the datasets and the weights of the confidence network are randomly initialised. A single training epoch consists of contrasting 10 patches of size $50\times 50$ (see Table~\ref{tab:hyperparam}) sampled from one random crop from each training image. At the end of each epoch, the confidence maps obtained from the confidence network are thresholded\footnotemark{} at $p=0.5$ and the segmentation performance is evaluated using the labels for the training images by computing the Dice score.  We limit the training of the models for a maximum of 300 epochs, as for both datasets we noticed that the learning plateaued around 200 epochs. After training for 300 epochs, the model at epoch with highest validation Dice score is selected for evaluation on the test set. These Dice scores are reported and discussed further in Section~\ref{sec:results}. Simple post-processing comprising two iterations of morphological opening and closing operations with radius 3 and 1, respectively, are performed on the thresholded segmentation masks; this primarily helps remove some high frequency noise in the predicted segmentation masks. 

\begin{figure}[t]
    \centering
    \includegraphics[width=0.499\textwidth]{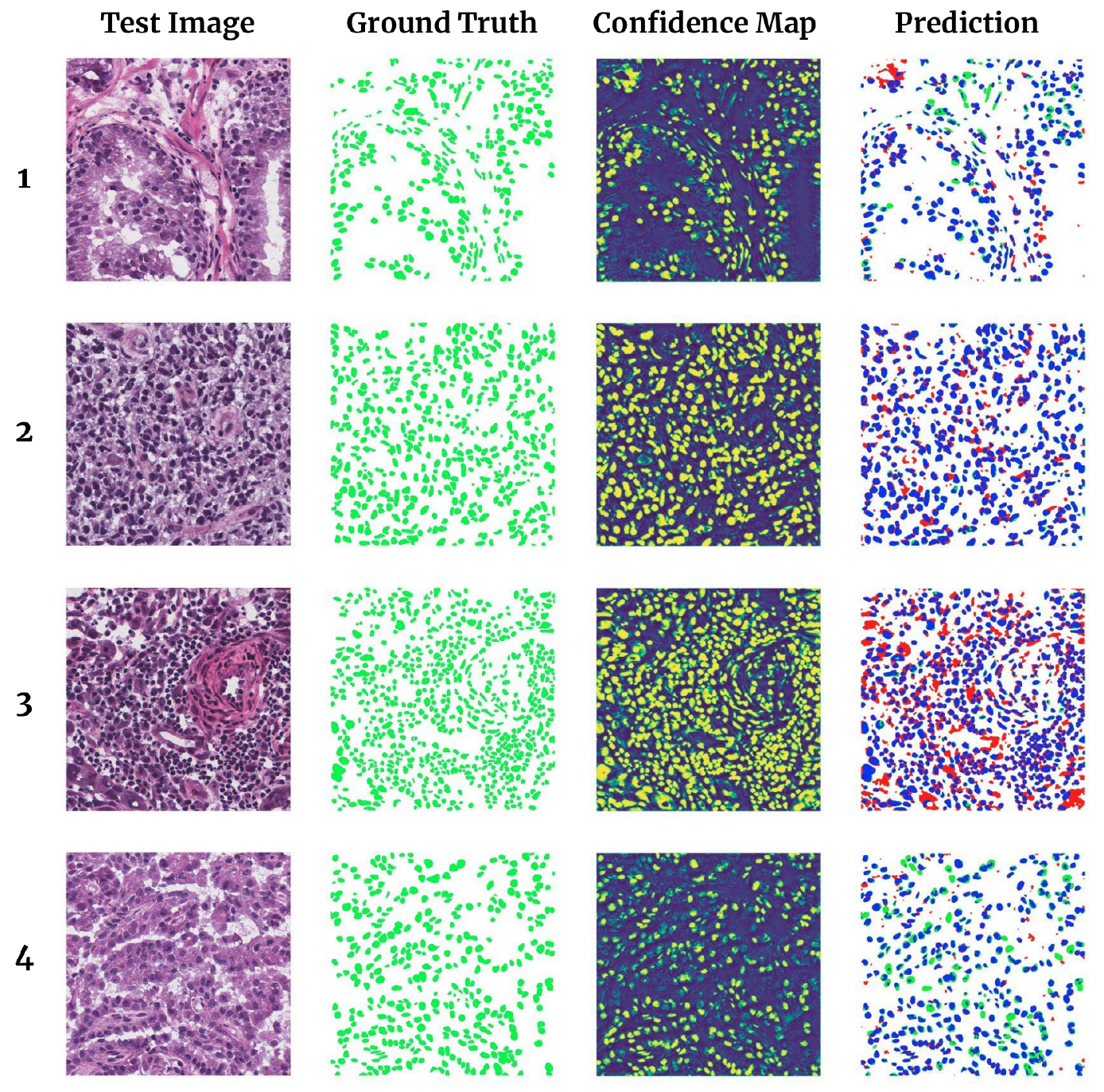}
    \caption{Four test set images from the MoNuSeg dataset. The rightmost column is colour-coded, where blue indicates correct segmentation (true positives), green indicates false negatives, red indicates false positives, and the white background corresponds to true negatives. The predictions are obtained from the confidence map by a $0.5$ threshold.}
    \label{fig:results}
\end{figure}

\footnotetext{The confidence maps, as illustrated in Figure \ref{fig:results}, are saturated to the extrema (close to 0 or 1); thus the segmentation performance might not be particularly sensitive to the specific choice of the threshold levels.}

The proposed CSL framework is trained 10 times with different initialisations on each dataset, in order to illustrate and quantify its inherent variance in segmentation performance.







\begin{table}[t]
\centering
\scriptsize
\begin{tabular}{@{}lllll@{}}
\toprule
{\bf Methods} & {\bf MoNuSeg} & {\bf CoNSeP} & {\bf \# Par.} & {\bf Time}
\\ \midrule
Int. Thresh.                                                                                   & 0.5823        & 0.5700                              & 1 &  $<$1m  \\ 
CNN                                                                                            &         0.7666                &       0.7375                          & 10.8k &  $\approx$ 10m   \\ \midrule
Scale Pretext 
& 0.6209        & 0.5139                      & 21.7M & $>$ 12h\footnotemark{}           \\ 
\multirow{3}{*}{\begin{tabular}[c]{@{}l@{}}{\bf Ours:} Max. \\ Mean $\pm$ s.d \\ 80 perc.  \end{tabular}} & 0.6797        & 0.6266               &              \\
& 0.62$\pm$0.05 & 0.52$\pm$0.06           &  10.8k & {$\approx$ 5m}  \\
& 0.6469 & 0.5448           & &                                                                            \\ \bottomrule
\end{tabular}
\caption{Dice scores on test sets of MoNuSeg and CoNSeP datasets for the different methods along with maximum, mean+sd and 80'th percentile scores for our method. Optimal intensity threshold method (Int. Thresh.), supervised CNN baseline, self-supervised method using scale prediction pretext task (Scale Pretext) and the proposed CSL framework. The number of trainable parameters and convergence time per method are also reported.}
\label{tab:results}
\end{table}

\footnotetext{This could be a large under-estimation of the actual convergence time based on in-house implementation as the actual run-time was not reported in \cite{sahasrabudhe2020self}.}

\section{Results} \label{sec:results}


Segmentation performance of our method and the baselines are reported in Table~\ref{tab:results} on both the datasets. The first two rows show the performance of two supervised methods and the third row shows the self-supervised Scale Pretext baseline method \cite{sahasrabudhe2020self}. We ran ten random repeats of our model for each dataset configuration and report the mean, maximum and the 80'th percentile Dice score over these runs. Dice score for the Scale Pretext method is obtained from the paper \cite{sahasrabudhe2020self}. In all cases, the supervised CNN method performs better than the self-supervised class of methods, which is to be expected, as no further fine-tuning of these methods with labels is performed.

\paragraph{MoNuSeg:} Our method outperforms the supervised optimal threshold method when trained on the MoNuSeg dataset using MSE-similarity. Qualitative results on four randomly selected test set images from the MoNuSeg dataset for the model trained on WSI data are shown in Figure~\ref{fig:results}. 

\paragraph{CoNSeP:} A similar performance trend is observed on the CoNSeP dataset, where we notice the mean Dice score of our method with the CE similarity measure performs better than the Scale Pretext method. However, the intensity thresholding method performs better than all self-supervised methods.  

The number of parameters of our method (10.8k) is several orders of magnitude fewer than Scale Pretext method (21M), and our method takes only a fraction of time until convergence (5m) compared to theirs ($>$12h) as shown in Table~\ref{tab:results}. This fast convergence time has implications in overcoming  the strong influence of the confidence maps obtained at initialisation (epoch-0) which can result in convergence to poor solutions. While some runs result in poor performance -- dragging the mean down -- most runs yield good performance. We capture this as the 80'th percentile performance over multiple runs, which is better than the mean performance. Thus running the model several times (which is still fast) will likely result in good performing solutions. As a standard practice, we therefore recommend training multiple repeats of our model on any dataset, and selecting the best performing one for inference.

\begin{figure*}[t]
    \centering
    \includegraphics[width=\textwidth]{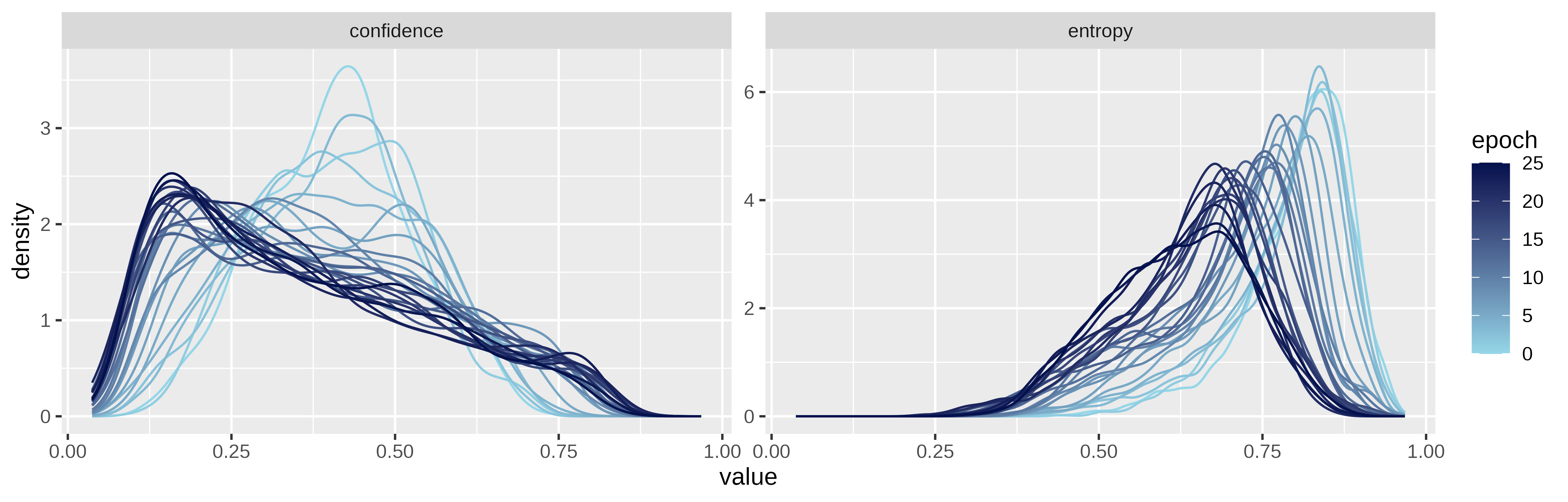}
    \caption{Evolution of confidence network inferences during training process. The two plots illustrate the distribution of average patch confidence and -entropy, respectively, stratified over epoch. This visualises how the distributions change during training; the broadening of the average patch confidence distribution corresponds to an increased information separation. Correspondingly, the average patch entropy decreases. We believe this to be an insight into the mechanism behind the proposed framework.}
    \label{fig:evolution}
\end{figure*}

\section{Discussions \& Conclusions}
\label{sec:disc}

\paragraphold{Confidence maps to segmentation masks:} The efficient CSL framework presented in this work outputs multiple confidence maps corresponding to objects believed to belong to the same class. Going from these confidence maps to the segmentation masks by definition requires expert input. In this work, we alleviate this by using the labels for model selection/validation. For other unlabelled datasets, some other form of validation would be required to align the concept of a biomedical image foreground to that of a confidence map of  a self-supervised framework. 
\paragraph{Self-supervised segmentation performance:} The main baseline we compared our method to is the self-supervised Scale Pretext method \cite{sahasrabudhe2020self}. The results of this method are not significantly different from ours. This is even with specific preprocessing (stain-normalisation) and elaborate post-processing followed in \cite{sahasrabudhe2020self}, which are altogether left out in our work. Further, compared to the Scale Pretext method ours is significantly simpler, both in terms of the model complexity, the elaborate regularisation of their objective function, and training time (5m versus 12h).



\paragraph{Mechanism:}
The aim of the proposed CSL framework is to increase the information separation between positive- and negative- patches. In Figure \ref{fig:evolution} the learning mechanism is illustrated, where the distribution of average patch confidence and entropy at each of the first 25 epochs of a training process are shown. Conceptually, and in an idealised setting, we would expect the distribution to evolve from being a unimodal distribution centered around $0.5$ to a bimodal distribution with peaks close to $0$ and $1$.  
While the average patch confidence in Figure \ref{fig:evolution} does not quite reach a bimodal distribution, it does arguably seem to evolve towards it. We conjecture that this training evolution plot illustrates the fundamental mechanism underlying the proposed CSL framework.

\paragraph{Limitations:} The main limitation in the proposed framework coincides with its simplicity --- there are no constraints encouraging the model to actually segment the desired objects of interest, potentially resulting in non-useful segmentation maps. Increasing $K$, the number of classes, may increase the chance of one of them being useful, though this comes at a considerable computational cost; hence the pragmatic choice of $K = 4$ throughout the paper. To further increase the chance of obtaining a useful model, we suggest a standard practice of training the model multiple times, and subsequently selecting the best performing one; the framework is quite sensitive to initialisation, resulting in considerable variance in performance. Fortunately, the 80-percentile scores in Table~\ref{tab:results} indicate that the framework often yields useful models, at least on these data sets. And training multiple times is tenable due to the efficiency of the framework.

Finally, we assume that the similarity measure plays an important role in model performance. We experimented with a few of these measures. However, thorough investigations of more refined similarity measures and their influence on multi-class segmentation remains future work.
\paragraph{Conclusions:} We presented a self-supervised framework for segmenting nuclei in histopathology data, which uses patch-based contrastive learning. We introduced a novel technique to mine for positive- and negative patches for contrasting, based on the average entropy of the confidence maps. This approach encourages the trainable confidence network to discern objects of different classes, increasing information separation. The resulting method with only 10.8k trainable parameters takes under 5 min. to converge, yielding useful segmentation masks. We foresee interesting research directions for this work that can make it better suited for diverse image data. 

\subsection*{Acknowledgements} The authors would like to thank members of Machine Learning Section, UCPH for useful feedback through the development of this work. NB thanks the Machine Learning Section, UCPH for their generous funding. RS is partly funded by the European Union's Horizon Europe research and innovation programme under grant agreements No. 101070284 and No. 101070408. 
\clearpage
\balance{
\bibliographystyle{abbrvnat}
\bibliography{references}}

\clearpage

\addcontentsline{toc}{section}{Appendix}
\appendix
\part{Appendix} 
\parttoc 

\section{Related Work}
\label{sec:related}
The high level ideas of contrastive learning and self-supervision have been widely studied in literature in recent years. Our work integrates some of these ideas into an efficient contrastive self-supervision framework that is able to handle complex biomedical imaging data. 

Many CSL methods are trained on large unlabelled datasets using an assortment of pretext tasks. These tedious training processes are time consuming, sometimes requiring several days of compute time~\cite{chen2020simple}. This could have a net negative impact due to the increase in their energy consumption compared to supervised learning, which might converge to useful solutions faster~\cite{anthony2020carbontracker,wu2022sustainable}. 

Due to the availability of several well-curated natural image datasets, bulk of the CSL methods are trained on these data~\cite{deng2009imagenet,lin2014microsoft}. The pre-training gains observed with self-supervised models on these natural image datasets do not immediately translate to more complex images that might not share features with natural images, such as the ones encountered in biomedical imaging~\cite{wen2021rethinking}. 
\\
{\bf Pretext task based self-supervision:} Self-supervised learning is a form of unsupervised learning which uses proxy or pretext tasks to obtain useful representations of the data. These pretext tasks can be simple data augmentation strategies such as rotation \cite{komodakis2018unsupervised} and colorisation \cite{larsson2016learning}. However, more complex pretext tasks such as patch context prediction \cite{doersch2015unsupervised} or solving image jigsaw puzzles \cite{noroozi2016unsupervised} have also shown to learn useful representations.
\\
{\bf Pre-training with self-supervised learning:} Most pretext based self-supervision methods are used as a pre-training step. These pre-trained models are further fine tuned, usually, using fewer labelled data than in fully supervised settings to achieve performance that is comparable to supervised methods, for instance to perform label efficient segmentation \cite{wang2021dense,zhao2021contrastive}. 
\\
{\bf Contrastive self-supervised learning:} Similarity based contrastive learning for obtaining discriminative representations has been studied in earlier works \cite{hadsell2006dimensionality}, but was formalised for self-supervised learning in the SimCLR framework \cite{chen2020simple}. The key challenge of contrastive self-supervision, however, is mining for hard examples. Simple data augmentation strategies such as the ones used in SimCLR \cite{chen2020simple} are not found to be very useful for discriminating complex features \cite{zhang2021unleashing}.
\\
{\bf Self-supervision and biomedical image segmentation:} Pretext tasks for improving nuclei segmentation from histopathology images have been formulated based on predicting the scaling factor of image regions \cite{sahasrabudhe2020self,xie2020instance}. This is based on the assumption that the background class in tissue imaging usually appears similar at different scales compared to nuclei which vary between scales. By using a pretext task of predicting the scale factor, the attention network is able to learn intermediate representations that are close to the nuclei segmentation mask. This scale based pretext task is very specific to the nuclei segmentation task, and might not generalise to other biomedical imaging data. 

\section{Sampling Methods}
\label{app:sampling}
Several sampling procedures including the entropy based sampling presented in Section \ref{sec:sampling}, have been tested during the development of this framework. We next present two most noteworthy alternatives as \emph{confidence based stochastic} sampling and \emph{top-$k$} sampling. 

\subsection{Confidence-based Sampling}
This sampling method is based on the probability distribution of average patch confidences of the patches in $S$. The probability of selecting a patch $(R, i) \in S$ for the set of positive patches is
\[ P(R) = \frac{A_k(R, i)}{\sum_{(T, j) \in S} A_k(T, j)}.  \]
A negative patch is sampled according to the probability distribution:
\begin{align*}
    N(R) &= \frac{1 - P(R)}{\sum_{T \in S} [1 - P(T)]} \\
    &= \frac{1 - P(R)}{|S| - \sum_{T \in S} P(T)} \\
    &= \frac{1 - P(R)}{|S| - 1.}
\end{align*}
Notice that the sets of positive- and negative- patches are not necessarily disjoint with this sampling procedure. For simplicity, the sampling is performed with replacement -- keeping all occurrences of selected patches -- turning the sampled sets into multi-sets.

\subsection{Top-$k$ sampling}
This sampling procedure partitions the set of candidate patches $S$ into positive- and negative- candidate sets, $P_c'$ and $N_c'$, respectively. Here, a patch $R$ is in $P_c'$ if it is among the $k$ patches with highest average patch confidence. Letting $k > n$ allows the introduction of some randomness by uniformly randomly sampling $n$ patches from $P_c'$ to the set of sampled positive patches and similarly for $N_c'$ into the set of sampled negative patches.


This sampling procedure performs reasonably well in our experiments for datasets with high contrast between features and does so in relatively few epochs. While the limited randomness leads to convergence in fewer epochs, it may be a limiting factor in harder datasets. 


\section{Similarity Measures}

Given the confidence map scaled images, several choices for the similarity measure can be explored:
\begin{enumerate}
\item {\bf Mean squared error (MSE)}: Given two patch locations, $(R_1, i)$ and $(R_2, j)$, with a bijective mapping, $g$, between their elements, the MSE similarity between them can be defined as
\begin{equation}
   \similarity\big((R_1, i), (R_2, j)\big) := - \frac{1}{|R_1|} \sum_{u \in R_1} \big[ F^k_i(u) - F^k_j(g(u)) \big]^2 
\end{equation}
where $i$ and $j$ may be the same image.


\item {\bf Cross entropy (CE)}: Similar to the MSE similarity, a mean binary CE similarity between two patches $(R_1, i)$ and $(R_2, j)$ with a bijective map $g$ may be defined as
\begin{align}
   \similarity\big((R_1, i), (R_2, j)\big) &:= \frac{1}{|R_1|} \sum_{u \in R_1} \big[ F_i^k(u) \log_2 F_j^k(g(u)) \nonumber \\ &+ (1 - F_i^k(u)) \log_2 (1 - F_j^k(g(u))) \big]. 
\end{align}

\item {\bf Feature network}: Another approach to comparing patches, which illustrates the abstractness of the requirements for the similarity measure and possibilities for generalisation, is to introduce a \emph{featuriser}, $\c F \colon \c X \to \bb R^\ell$, which maps patches into an $\ell$-dimensional feature space, again based on $F$. This featuriser can e.g.\ be a convolutional neural network. The similarity measure can then be
\begin{equation}
   \similarity\big((R_1, i), (R_2, j)\big) := \frac{\c F(F_i^k(R_1)) \cdot \c F(F_j^k(R_2))}{\|\c F(F_i^k(R_1))\| \, \|\c F(F_j^k(R_2))\|} 
\end{equation}
using the cosine similarity, which is comparable to what is done in SimCLR \cite{chen2020simple}.
\end{enumerate}

{\bf Choice of similarity measures:} The choice of similarity measure has an influence on our framework's performance, and different choices may suit different datasets better. The similarity measures tested in this work are all very simple; either directly pixel based like MSE or CE or grey-scale density based like KL-divergence or mutual information. Neither of them perform well in all cases, and we speculate that this framework has a bigger potential with more refined similarity measures. In the end, we only use MSE and CE across datasets, and report the best performance achieved with either of the two.

\section{Model Hyperparameters and Training}

\subsection{Patch shape}
For efficient processing on GPUs, the patches are chosen to be equally sized, rectangular, and aligned to the pixel grid.  Also requiring all input images $i = 1,\dots,M$ to be of the same size allows for further efficiency in implementation. The creation of the sets of candidate patches, $W_k$, as well as the calculation of $A_k(R, i)$ for all patches $(R, i) \in W_k$ across channels $k = 1,\dots,K$, can be performed in a single convolutional operation; using a non-trainable convolution kernel of the patch size with all values being the reciprocal of the kernel size, utilising striding to adjust the number of patches to create. For further improved data-parallelism, this operation can also be scaled across all channels simultaneously by adding a dimension to the kernel's weights and appropriately sparsifying the kernels with zero-entries.

\begin{figure}[t]
    \centering
    \includegraphics[width=0.49\textwidth]{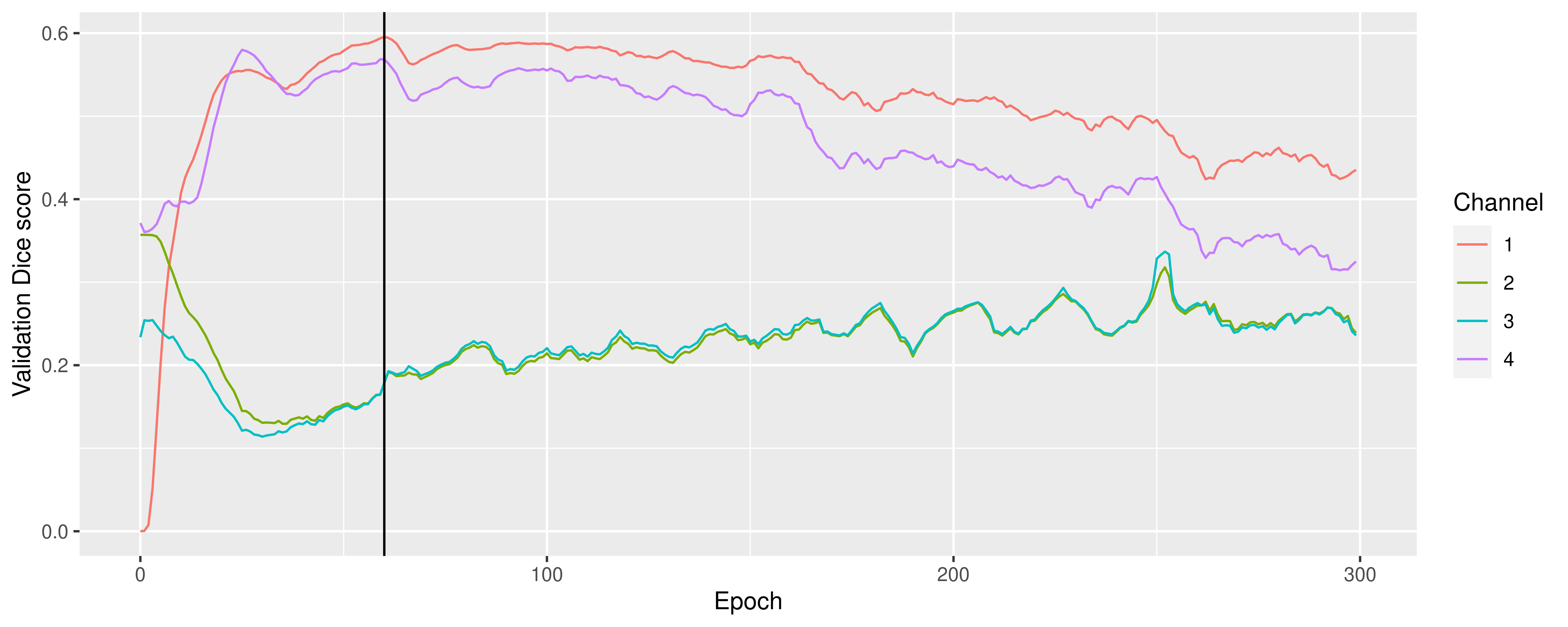}
    \caption{Validation Dice score during training, shown for all the $K = 4$ output channels. The vertical line depicts the epoch of the best validation Dice score, and hence the version of the model which was later evaluated on the test set. Notice that channels 2 and 3 are almost following each other, indicating a redundant output channel.}
    \label{fig:channels_111}
\end{figure}
\begin{figure}[t]
    \centering
    \includegraphics[width=0.49\textwidth]{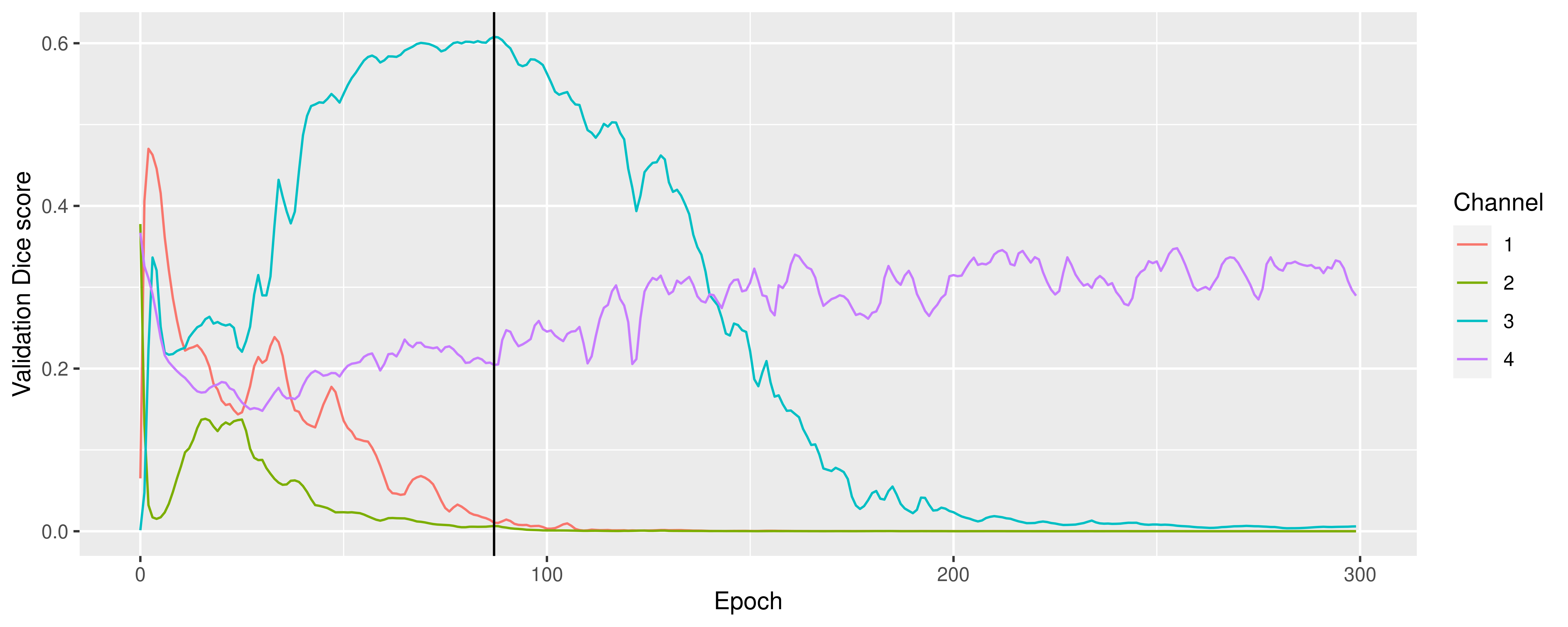}
    \caption{Validation Dice score during training, shown for all the $K = 4$ output channels. The vertical line depicts the epoch of the best validation Dice score, and hence the version of the model which was later evaluated on the test set. In this case, all four channels appear to be learning different features with channel 3 focusing on the foreground class.}
    \label{fig:channels_229}
\end{figure}

\subsection{Influence of Number of Output Channels}
\label{app:output}
Increasing the number of output classes/channels, $K$, increases the likelihood that at least one of the trained model's channels actually segments the desired features --- at least under the assumption that the different channels are trained towards recognising different features. The inter-channel contrastive loss is introduced to reward such class separation. 

Figures \ref{fig:channels_111} and \ref{fig:channels_229} illustrate the validation Dice score computed during training of two models. As is evident from Figure \ref{fig:channels_111}, not all $K = 4$ classes are properly separated. This can likely be explained by the intra-channel contrastive loss being a harsher penaliser than the inter-channel contrastive loss. This indicates that the inter-channel contrastive loss is not being enforced sufficiently. 

Figure \ref{fig:validation_111} illustrates predicted confidence maps and inferred segmentation maps from each of the $K = 4$ channels in the model presented in Figure \ref{fig:channels_111} predicts, at the epoch highlighted in that figure. Notice the same relation between channels as in Figure \ref{fig:channels_111}. Note that no morphological post-processing operations are performed, and that the segmentation performance is hence not representative of what is achieved in the final reporting. The model depicted here achieved a Dice score on the test set of 0.6797 after the post-processing.




\subsection{Confidence Network Architecture} \label{app:network_design}
The confidence network, $f_\theta$, in all configurations of our model are implemented as a fully convolutional neural network with residual connections, summarised as a sequence of
\begin{verbatim}
Conv2dBlock, s=1, p=d=1
ResidualBlock
    Conv2dBlock, s=1, p=d=1
    Conv2dBlock, s=1, p=d=1
ResidualBlock
    Conv2dBlock, s=1, p=d=2
    Conv2dBlock, s=1, p=d=2
ResidualBlock
    Conv2dBlock, s=1, p=d=3
    Conv2dBlock, s=1, p=d=3
ResidualBlock
    Conv2dBlock, s=1, p=d=5
    Conv2dBlock, s=1, p=d=5
ResidualBlock
    Conv2dBlock, s=1, p=d=10
    Conv2dBlock, s=1, p=d=10
ResidualBlock
    Conv2dBlock, s=1, p=d=20
    Conv2dBlock, s=1, p=d=20
Sigmoid
\end{verbatim}
where each \texttt{Conv2dBlock} consists of a sequence of a convolutional layer, batch normalisation, and ELU activation function, with \texttt{s} denoting the stride, \texttt{p} denoting the padding, and \texttt{d} denoting the dilation of the convolutional layer. All convolutional layers have kernel size $3 \times 3$. The first convolutional layer takes three input-channels (corresponding to colour images), and the last convolutional layer has $K$ output-channels. All other channel counts are 8. The \texttt{ResidualBlock}s computes the residual over the elements inside their indentation. This architecture is based on the attention network used in\cite{sahasrabudhe2020self}.

Notice that, in the context of recent large-scale deep learning, the confidence network with its 10.8K trainable parameters is tiny. This has implications on the training time and the compute resources (such as maximum GPU memory required), making the overall method more efficient.



\section{Datasets visualisation}
\begin{figure}[h]
    \centering
    \includegraphics[width=0.49\textwidth]{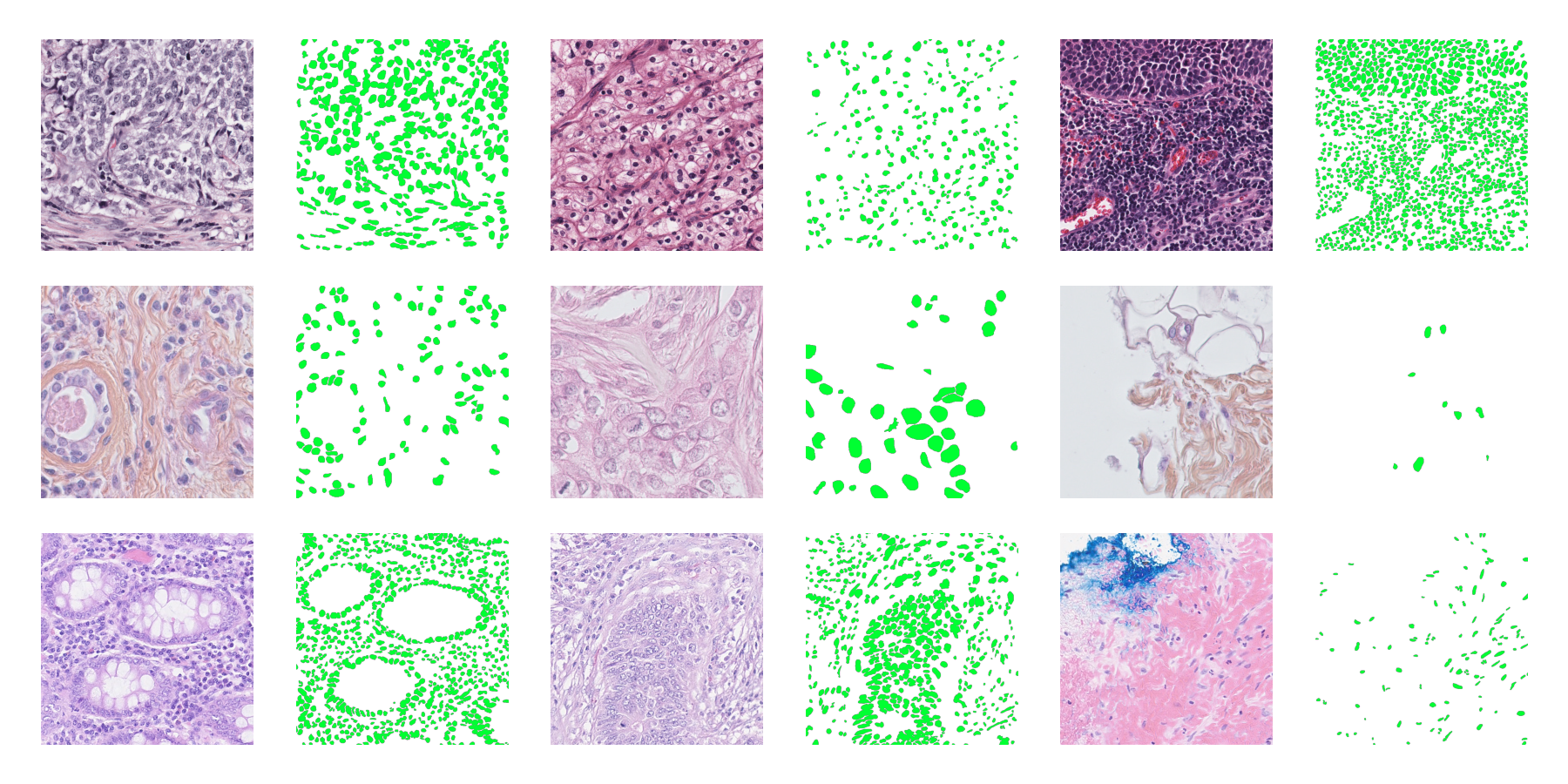}
    \caption{Three pairs of images and corresponding ground truth segmentation masks (in green) from each of the three datasets; MoNuSeg \cite{monuseg_data} (top row); TNBC \cite{naylor2018segmentation} (middle row); and CoNSeP \cite{graham2019hover} (bottom row). Notice differences in nuclei (foreground) counts, sizes, and colours/intensities.}
    \label{fig:dataset_examples}
\end{figure}

Figure \ref{fig:dataset_examples} illustrates the variance of data in the three tested datasets. Three examples are randomly chosen from each dataset and presented along with their ground truth segmentation masks.

\section{Additional Results}
\label{app:results}
\subsection{Qualitative Results}

\begin{figure*}[h]
    \centering
    \includegraphics[width=0.72\textwidth]{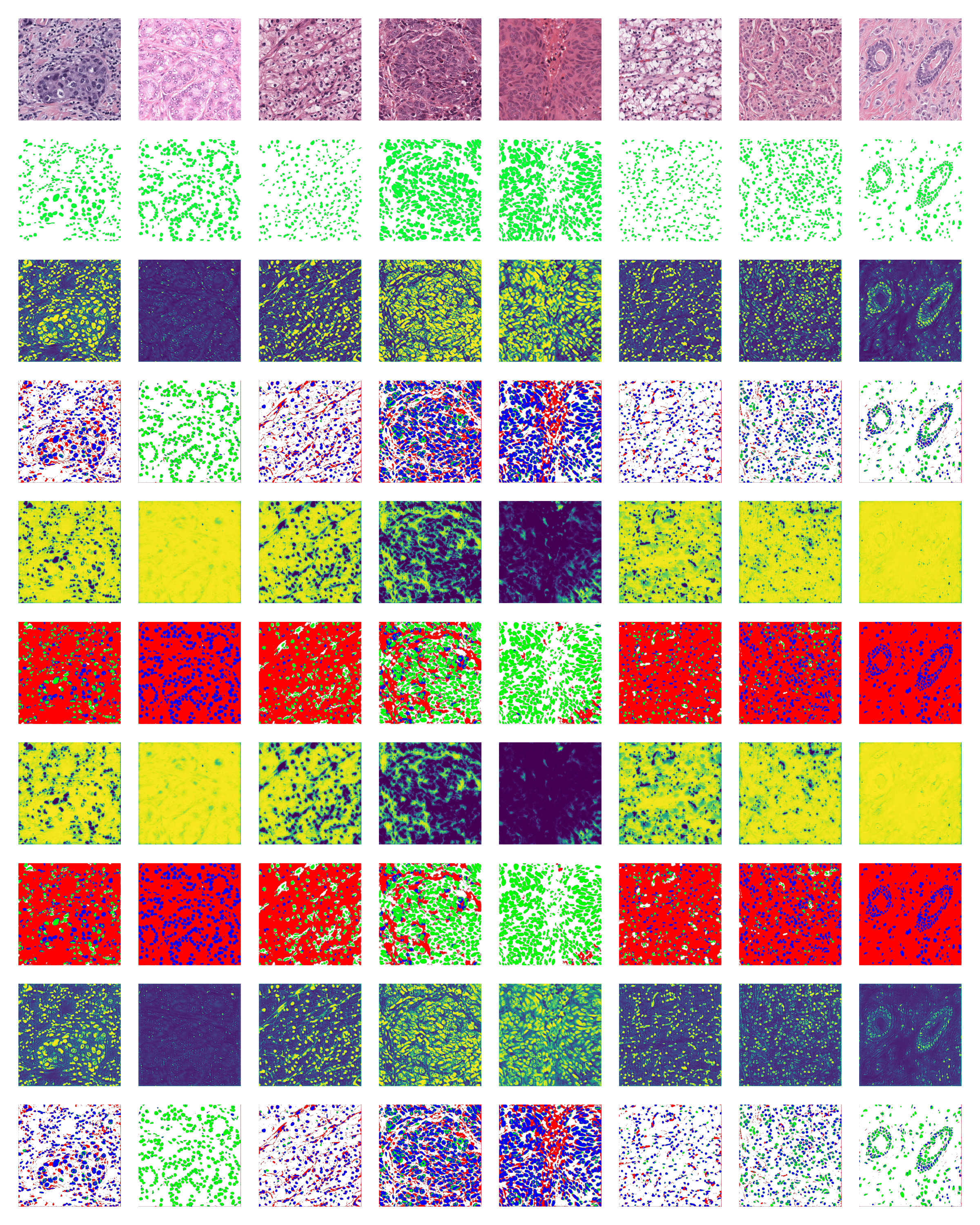}
    \caption{Predictions on validation set during training. Same model as Figure \ref{fig:channels_111}, and at the epoch with the best validation Dice score, which is depicted by the vertical line in that figure. Top row: Input images from validation set. Second row: Ground truth segmentation masks. In each of the $K = 4$ remaining pairs of rows: Upper row: Confidence map from that channel; Lower row: Predicted segmentation obtained by thresholding the confidence map, colour-coded as in Figure \ref{fig:results} where blue is true positive regions, green indicates false negatives, and red indicates false positives. No post-processing operations are performed.}
    \label{fig:validation_111}
\end{figure*}

Figure \ref{fig:validation_111} illustrates the prediction performance of a trained model using the proposed framework. This illustrates the necessity of multiple channels, as not all are useful for the intended task.
\subsection{Confusion matrices}
Table \ref{tab:cm:our} shows confusion matrices of proposed framework and the optimal threshold, respectively, on the MoNuSeg \cite{monuseg_data} dataset. Notice that the numbers correspond to the foreground constituting a minority of the total images.
\begin{table}[h]
\centering
\scriptsize
\begin{tabular}{@{}lcc|lcc@{}}
\toprule
\multicolumn{3}{l}{\bf Optimal Thresholding} & \multicolumn{3}{l}{\bf Proposed Method} \\ \midrule
Actual/Pred.        & P        & N       & Acutal/Pred.      & P      & N      \\ \midrule
P                   &   0.101       &  0.115       & P                 &    0.145    &   0.072     \\
N                   &   0.333       &  0.452       & N                 &   0.082     &   0.703     \\ \bottomrule
\end{tabular}
\caption{ Confusion matrices for the optimal thresholding method and the proposed framework on MoNuSeg}
\label{tab:cm:our}
\end{table}

\section{Experiments with Negative Results}

\subsection{TNBC dataset} In addition to the two histopathology datasets reported in the main paper, we also experimented on a third dataset: TNBC dataset~\cite{naylor2018segmentation}. This dataset consists of 50 images at $512 \times 512$ pixels, along with dense annotations of all nuclei. We split them into a training set of 28 images and testing set with the remaining 22 images; filenames ending 1 and 2 are used as test set, the remaining as train set. 

On the TNBC dataset, both versions (MSE, CE) of our method under-perform, as shown in Table~\ref{tab:results_app}. This could be due to the fact that the nuclei of interest have regions inside of them which are similar to the background class, according to the two similarity measures used. Another reason could also be the extreme class imbalance between the foreground and background pixels in this dataset, as seen in Figure~\ref{fig:dataset_examples}.

\begin{table*}[t]
\centering
\scriptsize
\begin{tabular}{@{}llllllll@{}}
\toprule
\multicolumn{1}{c}{\multirow{2}{*}{\bf Methods}}                           & \multicolumn{3}{c}{\bf MoNuSeg}                   & \multirow{2}{*}{\bf CoNSeP} & \multirow{2}{*}{\bf TNBC} & \multirow{2}{*}{\bf \# Param.} & \multirow{2}{*}{\bf Time}\\ \cmidrule(lr){2-4}
\multicolumn{1}{c}{}                                                                           & Subset        & Full          & WSI           &                         &                       \\ \midrule
Int. Thresh.                                                                                   & 0.5823        & 0.6205        & --            & 0.5700                  & 0.5590            & 1 &  $<$1m  \\ 
CNN                                                                                            &         0.7666      &   0.7867          &   --          &       0.7375            &     0.7121              & 10.8k &  $\approx$ 10m   \\ \midrule
Scale Pretext 
& 0.6209        & --            & 0.7477        & 0.5139                  & 0.5870    & 21.7M & $>$ 12h\footnotemark{}           \\ 
\midrule
\multirow{3}{*}{\begin{tabular}[c]{@{}l@{}}Ours (MSE): Max. \\ 80 perc. \\ Mean $\pm$ s.d\end{tabular}} & 0.6797        & 0.7088        & 0.7181        & 0.5507                  & 0.4324            \\
                                                                                               & 0.6469 & 0.6775 & 0.6911 & 0.4896           & 0.3884 &  &         \\
                                                                                               & 0.62$\pm$0.05 & 0.61$\pm$0.17 & 0.62$\pm$0.12 & 0.45$\pm$0.08           & 0.32$\pm$0.11 & 10.8k & {$\approx$ 5m}        \\ \midrule 
\multirow{3}{*}{\begin{tabular}[c]{@{}l@{}}Ours (CE): Max.  \\ 80 perc. \\ Mean $\pm$ s.d\end{tabular}}   & 0.6488        & 0.6176        & 0.6647        & 0.6266                  & 0.5101    \\ 
                                                                                               & 0.6354 & 0.6150 & 0.6432 & 0.5448           & 0.4833 & &         \\
                                                                                               & 0.59$\pm$0.05 & 0.52$\pm$0.18 & 0.63$\pm$0.02 & 0.52$\pm$0.06           & 0.37$\pm$0.20 & 10.8k& {$\approx$ 5m}         \\ \bottomrule
\end{tabular}
\caption{Dice scores on test sets of the three datasets for the different methods (maximum, 80'th percentile, and mean+sd). Optimal intensity threshold method (Int. Thresh.), supervised CNN baseline, self-supervised method using scale prediction pretext task (Scale Pretext) and the proposed CSL framework using two similarity measures: mean-squared error (MSE) and cross-entropy (CE). Convergence time and the number of trainable parameters per method are also reported.}
\label{tab:results_app}
\end{table*}
\subsection{Whole Slide Images in MoNuSeg}


The images in the MoNuSeg dataset originate from whole slide images (WSIs)~\footnote{Whole slide images are available at \url{https://portal.gdc.cancer.gov/}}, available at the TCGA portal. Inspired by~\cite{sahasrabudhe2020self} we fetch the 37 corresponding WSIs, from which the MoNuSeg training set originates, as a separate dataset, and use the 37 labelled MoNuSeg training images for validation of the segmentation performance and use 14 images as test set.

When trained on the additional WSI dataset, the Scale Pretext method appears to be benefit and outperforms our method~\ref{tab:results_app}. This indicates that additional training of our CSL method does not necessarily achieve performance gains. We speculate the simple FCNN (with 10.8k parameters) is unable to learn additional features from the new data. We hope to experiment with more complex confidence networks to test this hypothesis in future work.

\end{document}